\documentclass[letterpaper, 10 pt, conference]{ieeeconf}  

\IEEEoverridecommandlockouts                              

\usepackage{cite}
\usepackage{amsmath,amssymb,amsfonts}
\usepackage{algorithmic}
\usepackage{graphicx}
\usepackage{tabularx}
\usepackage{textcomp}
\usepackage{xcolor}
\usepackage{lscape}
\usepackage{makecell}
\usepackage{subcaption}
\usepackage{url}
\usepackage{color, colortbl}
\usepackage{multirow}
\usepackage{multirow}
\usepackage{tikz}

\definecolor{LightBlue}{RGB}{212, 250, 252} 
\definecolor{LightYellow}{RGB}{251, 255, 220}
\definecolor{LightGreen}{RGB}{190, 240, 203}

\newcommand\copyrighttext{%
  \footnotesize \textcopyright 2023 IEEE.  Personal use of this material is permitted.  Permission from IEEE must be obtained for all other uses, in any current or future media, including reprinting/republishing this material for advertising or promotional purposes, creating new collective works, for resale or redistribution to servers or lists, or reuse of any copyrighted component of this work in other works.}
\newcommand\copyrightnotice{%
\begin{tikzpicture}[remember picture,overlay]
\node[anchor=south,yshift=10pt] at (current page.south) {\fbox{\parbox{\dimexpr\textwidth-\fboxsep-\fboxrule\relax}{\copyrighttext}}};
\end{tikzpicture}%
}

\title{\LARGE \bf
Traffic Light Recognition using Convolutional Neural Networks: \\A Survey
}

\author{Svetlana Pavlitska$^{1,2}$, Nico Lambing$^{1}$, Ashok Kumar Bangaru$^{2}$ and J. Marius Zöllner$^{1,2}$
\thanks{$^{1}$ Department of Technical Cognitive Systems, FZI Research Center for Information Technology, Germany.
	{\tt\small \{surname\}@fzi.de}}%
\thanks{$^{2}$ Karlsruhe Institute of Technology (KIT), Germany.}
}

\begin{document}

\maketitle
\copyrightnotice
\thispagestyle{empty}
\pagestyle{empty}

\begin{abstract}

Real-time traffic light recognition is essential for autonomous driving. Yet, a cohesive overview of the underlying model architectures for this task is currently missing. In this work, we conduct a comprehensive survey and analysis of traffic light recognition methods that use convolutional neural networks (CNNs). We focus on two essential aspects: datasets and CNN architectures. Based on an underlying architecture, we cluster methods into three major groups: (1) modifications of generic object detectors which compensate for specific task characteristics, (2) multi-stage approaches involving both rule-based and CNN components, and (3) task-specific single-stage methods. We describe the most important works in each cluster, discuss the usage of the datasets, and identify research gaps. 

\end{abstract}

\section{INTRODUCTION}

\begin{figure}[t]
\centering
\begin{subfigure}[t]{0.9\linewidth}
    \includegraphics[width=1.0\textwidth]{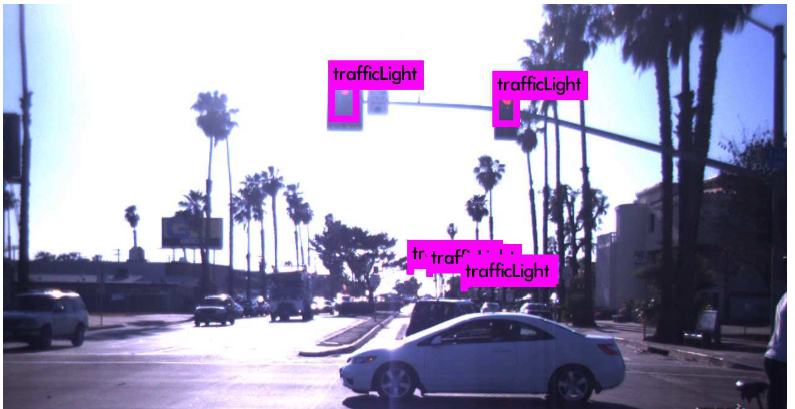}
  	\caption{TL detection~\cite{jensen2016vision}.}
\end{subfigure}

\begin{subfigure}[t]{0.9\linewidth}
    \includegraphics[width=1.0\textwidth]{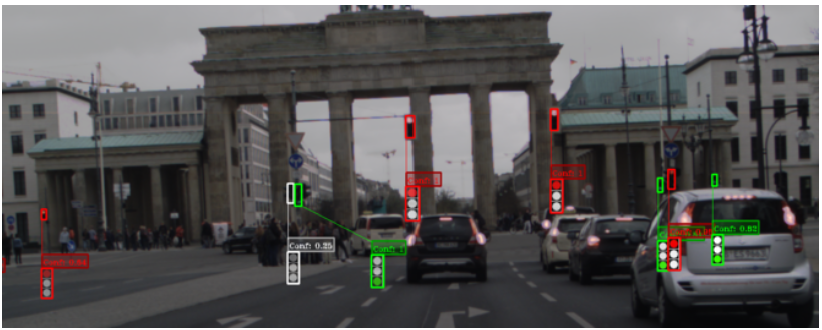}
  	\caption{TL detection + TL state classification~\cite{muller2018detecting}.}
\end{subfigure}

\begin{subfigure}[t]{0.9\linewidth}
    \includegraphics[width=1.0\textwidth]{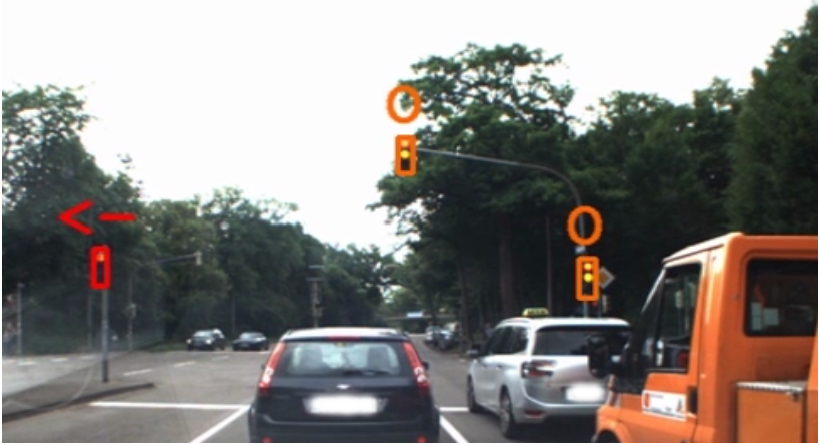}
  	\caption{TL detection + TL state and pictogram classification~\cite{weber2018hdtlr}.}
\end{subfigure}
    \caption{Examples of subtasks within the TLR task.}
    \label{fig:tlr}
\end{figure}

Detection and classification of traffic lights (TL) from camera images, also called \textit{traffic light recognition} (TLR), plays a pivotal role in enabling automated driving. It helps to maintain efficient and safe traffic flow management, reduce traffic congestion and minimize the risk of accidents. TLR as a task comprises \textit{traffic light detection}, which aims at localizing the traffic lights in the image, as well as \textit{classification of TL states} (colors) and \textit{pictograms} (arrows), as shown in Figure~\ref{fig:tlr}.  The development of convolutional neural networks (CNNs) has dramatically improved the accuracy of traffic light detection due to their ability to learn complex features from images. The effectiveness of CNN-based approaches depends on the choice of architecture and training data. 

In this work, we review and group existing CNN-based approaches for traffic light detection and classification. Unlike existing surveys on TLR, we focus on the choice of CNN architectures. 
Older surveys \cite{cabrera2015asurvey, jensen2016vision} focused more on classic image processing approaches since neural networks were only sporadically used then. To the best of our knowledge, the only concurrent modern work is that by Gautam et al.~\cite{gautam2023image}, which presents an in-depth overview but focuses on the whole pipeline. For the three main steps in the proposed pipeline (segmentation, feature extraction, and classification), Gautam et al. consider both classical computer vision approaches, like histograms of oriented gradients, and those using neural networks. In contrast, we focus on CNN model architectures and particularly on the modifications made to generic object detectors.

\section{Datasets for Traffic Light Recognition}



Since the appearance of traffic light signalling devices varies over different countries, a number of TLR benchmarks has been released. We provide an  overview of publicly available datasets in Table~\ref{tab:datasets}. We also refer to the journal paper by Jensen et al.~\cite{jensen2016vision}, which gives a comprehensive overview of datasets published before 2016.


\begin{table*}[ht]
    \caption{Comparison of TLR datasets ($*$ -- the number of classes in the test subset)}
    \centering
    \resizebox{1.0\textwidth}{!}{
        \begin{tabular}[t]{|rcc|cccccccccc|}
        \hline
        \textbf{Dataset}  &  \textbf{Year}  &  \textbf{Ref.}  &  \textbf{\makecell{Number of\\ images}}  &  \textbf{Resolution}  &  \textbf{Depth [bit]}  &  \textbf{\makecell{Frame \\Rate [Hz]}}  &  \textbf{\makecell{Number of\\ annotations}}  &  \textbf{\makecell{Disparity\\Data}}  &  \textbf{\makecell{Pictograms}}  & \textbf{Classes}  &  \textbf{Country}  &  \textbf{License}\\
        \hline
        LaRa & 2015 & \cite{LARA} & 11,179 & 640$\times$480 & 8 & 25 & 9,168 &   &  & 4 & France & N/A\\
        
        LISA & 2016 & \cite{jensen2016vision} & 43,007 & 1280$\times$960 & 8 & 16 & 119,231 & $\checkmark$ & $\checkmark$& 7 & USA & CC BY-NC-SA 4.0\\
        
        WPI & 2016 & \cite{WPI} & 3,456 & 1920$\times$1080 & N/A & N/A & 6766 &   & $\checkmark$& 21, 2 & USA & N/A\\
        
        BSTLD & 2017 & \cite{behrendt2017deep} & 13,427 & 1280$\times$720 & 8, 12 & 15 & 24,242 & $\checkmark$ &  $\checkmark$& 15, 4$^*$ & USA & MIT\\
        
        DriveU v1.0 & 2018 & \cite{fregin2018driveu} & 40,979 & 2048$\times$1024 & 8, 16 & 15 & 232,039 & $\checkmark$ & $\checkmark$& 423 & Germany & Academic\\
        
        
        DriveU v2.0 & 2021 & \cite{fregin2018driveu} & 40,979 & 2048$\times$1024 & 8, 16 & 15 & 292,245 & $\checkmark$ & $\checkmark$& 620 & Germany & Academic\\
        
        
        Cityscapes TL++ & 2022 & \cite{janosovits2022cityscapes} & 5,000 & 2048$\times$1024 & 16 & 17 & N/A & $\checkmark$ & & 6 & Germany & LGPL-2.1\\
        
        S$^2$TLD & 2022 & \cite{yang2022scrdet++} & 5,786 & 1080$\times$1920, 720$\times$1280 & N/A  & N/A  &  14,130 &     &  & 5 & China & MIT\\
        \hline
        \end{tabular}
        \label{tab:datasets}
    }
\end{table*}

\textbf{La Route Automatis\'ee (LaRa) dataset}~\cite{LARA} was one of the first publicly available datasets published in 2015 by a French joint research unit La Route Automatis\'ee . It contains over 11,000 images and 9,000 annotations recorded as a 25 Hz video during about a 9-minute long ride in Paris. The images have a relatively low resolution of $640 \times 480$ pixels.  All labels were annotated manually as bounding boxes (BBoxes) with object IDs for tracking evaluation. The TL state was labeled as \textit{green, orange, red}, or \textit{ambiguous}. Furthermore, each image was annotated with a sequence ID and timestamp. 

\textbf{LISA Traffic Light Dataset}~\cite{jensen2016vision} is a comprehensive dataset that contains over 40,000 images, originating from the Vision for Intelligent Vehicles and Application (VIVA) challenge, which included  the traffic light detection benchmark. Therefore, the dataset itself is sometimes also referred to as the \textit{VIVA dataset}. The data was captured as a 10 Hz video using a stereo camera with an image resolution of $1280 \times 960$ pixels and a horizontal field of view of approximately 43°. 
Additionally, depth disparity maps for each image are provided. The dataset consists of a training subset and a test subset, the latter is kept private to serve as the basis for benchmarking.  All labels were annotated manually and are provided as pixel-level binary masks and BBoxes. TL states are encoded using seven classes: \textit{go, go forward, go left, warning, warning left, stop, stop left}. The LISA dataset covers several USA cities (San Francisco, Berkeley, and Chicago) under different lighting and weather conditions.


\textbf{Bosch Small Traffic Lights Dataset (BSTLD)}~\cite{behrendt2017deep} was 
recorded along the El Camino Real in California's San Francisco Bay Area using a stereo camera. 
The dataset includes the corresponding disparity maps. All labels were annotated manually utilizing 15 classes to describe the color and pictogram of the TLs. The labels are provided as pixel-wise binary masks and BBoxes and are split into a training set and a test set of nearly equal size. Although the training data is labeled with a full set of 15 classes, the test data includes only four classes (\textit{red, yellow, green, off}).


\textbf{DriveU Traffic Light Dataset (DTLD) v1.0}~\cite{fregin2018driveu} was published in 2016 by the Intelligent User Interfaces (IUI) group at the University of Ulm in Germany. It has a number of images comparable to LISA but exceeds all other datasets in terms of the number of annotations (more than 230,000). Images with a resolution of $2048 \times 1024$ pixels were recorded by a stereo camera with a frame rate of 15 Hz. The dataset includes the corresponding disparity maps. 

The DTLD v2.0 dataset was released in 2021 as an extension of the DTLD v1.0. It contains images of the same resolution and frame rate but covers a broader range of traffic scenarios, such as roundabouts and T-junctions. Both datasets were annotated with BBoxes using manual and semi-automatic methods. The manual annotation was performed by human annotators, who labeled the TLs with pixel-level accuracy. The semi-automatic annotation was performed using a deep neural network trained to detect and classify TLs in the images.

Both datasets provide a comprehensive set of labels, arranged into the following groups: the orientation (\textit{front, back, left, right}), relevance/occlusion, orientation (\textit{horizontal, vertical}), the number of lamps, state (\textit{red, yellow, green, red-yellow, off}), and pictogram (\textit{circle, arrow left, pedestrian}, etc.). Because of the large number of possible combinations of these tags, the resulting number of unique labels exceeds that of any other dataset. DTLD v1.0 and v2.0 were collected from eleven German cities, including urban and suburban environments, to provide diverse TL scenarios. 


Furthermore, a number of \textbf{other public datasets} either include labeled TL states or have been extended to include them. However, they lack additional attributes such as orientation, pictogram, and relevance information, which are necessary to utilize the detected TLs for autonomous driving. Examples of the datasets extended with TL states include COCO Traffic~\cite{COCO}, where TL states were annotated in the images from the COCO~\cite{lin2014microsoft} dataset, as well as Cityscapes TL++ dataset~\cite{janosovits2022cityscapes} containing images with fine annotations from the Cityscapes~\cite{Cordts2016Cityscapes} dataset with additional TL labels for four attributes: type (\textit{car, pedestrian, bicycle, train, unknown}), relevant (\textit{yes, no}), visible  (\textit{yes, no}), and state (\textit{red, red-yellow, yellow, green, off, unknown}). Other datasets containing only TL state labels are the Roboflow Self-Driving Car dataset~\cite{Roboflow}, a modified version of the Udacity Self-Driving Car Dataset~\cite{Udacity}, Waymo Open Dataset~\cite{Waymo}, WPI~\cite{WPI}, BDD100K~\cite{BDD100k}, and ApolloScape~\cite{ApolloScape} datasets. 

\section{Overview of Architectures for Traffic Light Recognition}

Compared to the generic object detection task, specific challenges in TLR include small object size, sparse structure, and high variability of the background. Various works have proposed different methods to approach these issues. We cluster them into three groups: (1) \textbf{modifications} of generic object detectors, (2) \textbf{multi-stage approaches}, which perform TL localization and TL state/pictogram classification in separate steps, and (3) \textbf{task-specific single-stage approaches}, which perform TLR within a single network.  

Table~\ref{tab:approaches} summarizes existing work on CNN-based TLR approaches. In the following, we give an overview of the most important works in each group. Please note that we have deliberately omitted approaches involving only TL classification, without previous detection step (e.g., Gautam and Kumar~\cite{gautam2022automatic}), as they are unrealistic for the deployment.

{\setlength{\fboxsep}{0pt}
\begin{table*}[t]
\caption{Overview of TLR approaches: \colorbox{LightGreen}{modifications of generic detectors}, \colorbox{LightBlue}{multi-stage approaches}, \colorbox{LightYellow}{task-specific single-stage approaches}. Backbone architecture is stated in parentheses. Inference speed and accuracy are mentioned if provided in the corresponding publication. FPS were converted to ms for better compatibility.}
\label{tab:approaches}
\centering
  \resizebox{1.0\textwidth}{!}{
    \begin{tabular}[t]{|lcc|llllccc|}
    \hline
    \textbf{Author}  &  \textbf{Year}  &  \textbf{Ref.}  &  \textbf{\makecell{Approach}}  &  \textbf{Dataset} & \textbf{\makecell{Inference\\ speed}} & \textbf{\makecell{Accuracy}}   & \textbf{\makecell{TL \\states\\classified}}  &  \textbf{\makecell{TL \\pictograms\\classified}}  &  \textbf{\makecell{Source\\code}}   \\  \hline
    \rowcolor{LightYellow}
    John et al. & 2014 & \cite{john2014traffic,john2015saliency} &  CNN similar to LeNet & \colorbox{LightYellow}{\makecell[l]{Private (USA, \\Japan, France)}}  & 10 ms & Accuracy: 96.25-99.4\% & $\checkmark$ &   &    \\ \hline
    
    \rowcolor{LightYellow}
    Weber et al. & 2016 & \cite{weber2016deeptlr} & DeepTLR (single CNN) &  Private (Germany) & 
    30-77 ms& \colorbox{LightYellow}{\makecell[l]{F1: 93.5\%}}  &  $\checkmark$ &   &     \\ \hline
    
    \rowcolor{LightBlue}
    Behrendt et al. & 2017 & \cite{behrendt2017deep} &  \colorbox{LightBlue}{\makecell[l]{Detection: YOLOv1,\\Classification: 6-layer CNN}} & BSTLD  & 
    67-100 ms& F1: $\sim$80\% &$\checkmark$ &   &     \\ \hline

    \rowcolor{LightGreen}
    Jensen et al. & 2017 & \cite{jensen2017evaluating} &  Modified YOLOv2  & LISA, LaRa & N/A & AUC: 90.49\% (LISA) &   &   &     \\ \hline

    \rowcolor{LightYellow}
    Weber et al. & 2018 & \cite{weber2018hdtlr} & HDTLR (single CNN) & \colorbox{LightYellow}{\makecell[l]{BSTLD,\\Private (Germany)}} &  
    83 ms& \colorbox{LightYellow}{\makecell[l]{F1: 85.8\% (BSTLD),\\F1: 88.8\% (private)}} &$\checkmark$ & $\checkmark$ &     \\\hline

    \rowcolor{LightGreen}
    Müller and Dietmayer & 2018 & \cite{muller2018detecting} & \colorbox{LightGreen}{\makecell[l]{Modified SSD\\(Inception-v3)}} & DTLD & 
    100 ms& Recall: 95\% &$\checkmark$ &   & $\checkmark$ \\ \hline 

    \rowcolor{LightGreen}
    Pon et al. & 2018 & \cite{pon2018hierarchical} &  Faster R-CNN (ResNet-50) & BSTLD & 15 ms & mAP: 53\% & $\checkmark$ &   &     \\ \hline

    \rowcolor{LightGreen}
    Bach et al. & 2018 & \cite{bach2018deep} &  Modified Faster R-CNN (ResNet-50) & DTLD  & N/A & mAP: 83\% &$\checkmark$ & $\checkmark$ &     \\ \hline

    \rowcolor{LightBlue}
    Kim et al. & 2018 & \cite{kim2018efficient} &  \colorbox{LightBlue}{\makecell[l]{Color space transformation + \\an ensemble of 3 networks: \\ Faster R-CNN (Inception-ResNet-v2\\ or ResNet-101) or R-FCN (ResNet-101)}} & BSTLD & N/A & mAP: 38.48\%& $\checkmark$ &   &     \\ \hline

    \rowcolor{LightBlue}
    Lu et al. & 2018 & \cite{lu2018traffic} &  \colorbox{LightBlue}{\makecell[l]{Visual attention proposal + detection,\\both based on Faster R-CNN}} & \colorbox{LightBlue}{\makecell[l]{LISA,\\Private (China)}} & N/A & mAP: 91.1\% (LISA)&$\checkmark$ & $\checkmark$ &    \\ \hline
    \rowcolor{LightBlue}
    Wang et al. & 2018 & \cite{wang2018traffic} & \colorbox{LightBlue}{\makecell[l]{ROI detection: HDR-based\\saliency map filtering,\\Classification: AlexNet}} & Private (Singapore) & 35 ms & mAP: 98.9\% &$\checkmark$ & $\checkmark$ &    \\ \hline
    
    \rowcolor{LightBlue}
    Kim et al. & 2018 & \cite{kim2018deep} &  \colorbox{LightBlue}{\makecell[l]{SSD for coarse-grained \\detection + spatiotemporal \\refinement}}  & Private (USA) & N/A & F1: 10.05\% - 69.68\% & $\checkmark$ & $\checkmark$ &   \\ \hline

    \rowcolor{LightBlue}
    Wang et al. & 2018 & \cite{wang2018method} &  \colorbox{LightBlue}{\makecell[l]{Detection: YOLOv3,\\Classification: 4-layer CNN}}  & BDD110K  & 35 ms & Accuracy: 98\% &$\checkmark$ &   &  \\ \hline

    \rowcolor{LightBlue}
    Yudin et al. & 2018 & \cite{yudin2018usage} &  \colorbox{LightBlue}{\makecell[l]{Detection: fully-connected network\\ + binarization + clustering}} & \colorbox{LightBlue}{\makecell[l]{Nexar TLR \\ Challenge}} & 63 ms & \colorbox{LightBlue}{\makecell[l]{Recall: 94.37\%,\\Precision: 43.23\%}} &  &    & $\checkmark$ \\ \hline 
    
    \rowcolor{LightGreen}
    Han et al. & 2019 & \cite{han2019real} & \colorbox{LightGreen}{\makecell[l]{Modified Faster R-CNN (VGG16)}} & Private (China) & N/A & mAP: 49.26\% &  &   &    \\ \hline

    \rowcolor{LightBlue}
    Possatti et al. & 2019 & \cite{possatti2019traffic} &  YOLOv3 + prior maps &  \colorbox{LightBlue}{\makecell[l]{DTLD, LISA, \\Private (Brazil)}}  & 
    48 ms& \colorbox{LightBlue}{\makecell[l]{mAP: 85.62\% (DTLD)\\ mAP: 50.59\% (LISA)}}  &$\checkmark$ &   & $\checkmark$  \\ \hline 

    \rowcolor{LightGreen}
    Ennahhal et al. & 2019 & \cite{ennahhal2019real} & \colorbox{LightGreen}{\makecell[l]{Faster R-CNN (ResNet-101,\\ Inception V2), R-FCN (ResNet-101), \\ SSD (MobileNet)}} & BSTLD, LISA & 
    200-333 ms& mAP: 79.01\%&$\checkmark$ &   &   \\ \hline

    \rowcolor{LightBlue}
    Gupta and Choudhary & 2019 & \cite{gupta2019framework} &  \colorbox{LightBlue}{\makecell[l]{Detection: Faster R-CNN (VGG16),\\Classification: Grassmann manifold\\learning}} & \colorbox{LightBlue}{\makecell[l]{BSTLD, LaRa,\\LISA, WPI}}  & 31 ms & \colorbox{LightBlue}{\makecell[l]{Accuracy: 98.80\%}} &$\checkmark$ & $\checkmark$ &   \\ \hline

    \rowcolor{LightGreen}
    Du et al. & 2019 & \cite{du2019real} &  YOLO3 & Private (China) & 106 ms & mAP: 96.18\% &  &   &  \\ \hline

    \rowcolor{LightBlue}
    Yeh et al. & 2019 & \cite{yeh2019traffic,yeh2021traffic} &  \colorbox{LightBlue}{\makecell[l]{Detection: YOLOv3,\\TL state classification: YOLOv3-tiny\\Pictogram classification: LeNet}} & \colorbox{LightBlue}{\makecell[l]{LISA,\\Private (Taiwan)}}  & 31-52 ms & mAP: 66\% (LISA) &$\checkmark$ & $\checkmark$ &  \\ \hline

    \rowcolor{LightBlue}
    Kim et al. & 2019 & \cite{kim2019traffic} &  \colorbox{LightBlue}{\makecell[l]{Detection: ENet-based network,\\Classification: LeNet-based CNN}} & BSTLD  & 
    34 ms& F1: 95.10\%&$\checkmark$ &   &  \\ \hline

    \rowcolor{LightGreen}
    Aneesh et al. & 2019 & \cite{aneesh2019real} &  RetinaNet (ResNet-50)  & BSTLD & 108 ms& mAP: 38.07\% & $\checkmark$ &   &   \\ \hline

    \rowcolor{LightBlue}
    Vishal et al. & 2019 & \cite{vishal2019traffic} &  \colorbox{LightBlue}{\makecell[l]{Detection: YOLO,\\Classification: color-based area\\ extraction and SVM}}  & BSTLD & 
    143 ms& F1: 94\% & $\checkmark$ &   &   \\ \hline

    \rowcolor{LightBlue}
    Cai et al. & 2019 & \cite{cai2019deltr} &  \colorbox{LightBlue}{\makecell[l]{Detection: SSDLite (MobileNetV2),\\Classification: 3-layer CNN}}  & BDD100K & \colorbox{LightBlue}{\makecell[l]{100 ms + 0.7ms}} & \colorbox{LightBlue}{\makecell[l]{Recall:95.3\%\\Precision:95.2\%\\mAP:33.84\%}} & $\checkmark$ &   &   \\ \hline

    \rowcolor{LightGreen}
    Janahiraman et al. & 2019 & \cite{janahiraman2019traffic} &  \colorbox{LightGreen}{\makecell[l]{SSD (MobileNetV2), \\ Faster R-CNN (Inception-v2)}} & Private (Malaysia) & N/A & mAP: 97.02\% &  &   &   \\ \hline

    \rowcolor{LightBlue}
    Ouyang et al. & 2020 & \cite{ouyang2020deep} &  \colorbox{LightBlue}{\makecell[l]{Detection: heuristic ROI detector,\\Classification: 18-layer CNN}}   & \colorbox{LightBlue}{\makecell[l]{WPI, LISA,\\Private (China)}}  & 
    \colorbox{LightBlue}{\makecell[l]{53 ms (WPI)\\ 43 ms (LISA)}}
 & Accuracy 99.7\%& $\checkmark$ &   &    \\ \hline

    \rowcolor{LightBlue}
    Tran et al. & 2020 & \cite{tran2020accurate} & YOLOv4 + color-based post-processing  & Private (South Korea) & 
    33 ms& Accuracy: 95\% &$\checkmark$ &   &   \\ \hline

    \rowcolor{LightBlue}
    Nguyen et al. & 2020 & \cite{nguyen2020robust} &  YOLOv3 + rule-based validation   & CCD~\cite{nguyen2019realtime} & N/A & Detection rate: 80\%& $\checkmark$ &   &   \\ \hline


    \rowcolor{LightBlue}
    Gao et al. & 2020 & \cite{gao2020hybrid} &  \colorbox{LightBlue}{\makecell[l]{Detection: ROI detector using \\HSV color space,\\Classification: AlexNet}} & LISA, LaRa  & 13-21 ms& Accuracy: 85.30\%& $\checkmark$ &   &  \\ \hline

    \rowcolor{LightBlue}
    Vitas et al. & 2020 & \cite{vitas2020traffic} & \colorbox{LightBlue}{\makecell[l]{Detection: adaptive thresholding,\\Classification: 3-layer CNN}} & LISA & N/A & Detection rate: 89.60\%& $\checkmark$ &   &     \\ \hline    

    \rowcolor{LightGreen}
    Gokul et al. & 2020 & \cite{gokul2020comparative} &  Faster R-CNN, YOLOv2, YOLOv3 & BSTLD & 159 ms& mAP: 48.64\%& $\checkmark$ &   &    \\ \hline
 
    \rowcolor{LightGreen}
    Abraham et al.  & 2021 & \cite{abraham2021traffic} & Modified YOLOv4-CSP &  Private (Indonesia)  & 
    34 ms & mAP: 79.77\% & $\checkmark$ &   &    \\ \hline


    \rowcolor{LightGreen}
    Yan et al. & 2021 & \cite{yan2021end} &  YOLOv5    & BDD100K & 
    7 ms& AP: 63.3\%& $\checkmark$ &   &    \\ \hline

    \rowcolor{LightGreen}
    Xiang et al. & 2021 & \cite{xiang2021real} &  Modified YOLOv3  & \colorbox{LightGreen}{\makecell[l]{LaRa,\\CQTLD (China)}} & 
    18 ms& mAP: 98.76\% & $\checkmark$ &   &    \\ \hline

    \rowcolor{LightGreen}
    Naimi et al. & 2021 & \cite{naimi2021fast} &  Modified SSD (MobileNetC2) & Private (Japan) & 443 ms& mAP: 73.8\%& $\checkmark$ &   &     \\ \hline 

    \rowcolor{LightGreen}
    Wang et al.  &  2022  & \cite{wang2022traffic}  & Modified YOLOv4 (CSPDarknet-53)  &   LaRa, LISA & \colorbox{LightGreen}{\makecell[l]{34 ms (LISA)\\40 ms (LaRa)}} & \colorbox{LightGreen}{\makecell[l]{mAP: 82.15\% (LISA)\\ mAP: 79.97\& (LaRa)}} & $\checkmark$ &   &     \\ \hline


    \rowcolor{LightGreen}
    Zhao et al. & 2022 & \cite{zhao2022study} &  YOLOv4 (ShuffleNetv2) & \colorbox{LightGreen}{\makecell[l]{S$^2$TLD,\\Private (China)}}   & 
    31 ms& \colorbox{LightGreen}{\makecell[l]{mAP: 71.24\% (S$^2$TLD)\\mAP: 62.12\% (private)}} & $\checkmark$ &   &   \\ \hline

    \rowcolor{LightBlue}
    Bali et al. & 2022 & \cite{bali2022lightweight} &  \colorbox{LightBlue}{\makecell[l]{Feature extraction: SqueezeNet,\\Classification: YOLOv2}}  & LaRa  & N/A & mAP: 84\%&$\checkmark$ &   &    \\\hline

    \rowcolor{LightYellow}
    Wang et al. & 2022 & \cite{wang2022simultaneous} & \colorbox{LightYellow}{\makecell[l]{A CNN and integrated\\ channel feature tracking}} & \colorbox{LightYellow}{\makecell[l]{BDD100K,\\ private (China)}} & 
    48 ms& F1: 84.7\% & $\checkmark$ &   &    \\\hline

    \rowcolor{LightBlue}
    Jayasinghe et al. & 2022 & \cite{jayasinghe2022towards} &  \colorbox{LightBlue}{\makecell[l]{Detection: Faster R-CNN (ResNet-50)\\ or SSD (MobileNet-v2),\\Classification: ResNet-18}} & Private (Sri Lanka) & 
    16 ms& F1: 92.14\% & $\checkmark$ &   &    \\ \hline
    
    \rowcolor{LightGreen}
    Mostafa et al. & 2022 & \cite{mostafa2022yolo} &  YOLOv4  & \colorbox{LightGreen}{\makecell[l]{LISA,\\Private (Egypt)}} & N/A & mAP: 92.16\% (LISA) & $\checkmark$ &   &    \\ \hline

    \rowcolor{LightBlue}
    Lin et al. & 2022 & \cite{lin2022two} &  \colorbox{LightBlue}{\makecell[l]{Detection: Faster R-CNN (ResNet-50),\\Classification: VGG16}}  &  Private (Taiwan) & 680 ms& mAP: 81.9\% - 86.4\%& $\checkmark$ &   &   \\ \hline


    \rowcolor{LightGreen}
    DeRong and ZhongMei & 2023 & \cite{derong2023remote} &  YOLOv5, YOLOv5+DeepSort & Private (China) & N/A & N/A & $\checkmark$ &   &    \\ \hline

    \rowcolor{LightGreen}
    Liu and Li & 2023 & \cite{liu2023traffic} &  \colorbox{LightGreen}{\makecell[l]{YOLOv5 (custom backbone)}}  & BSTLD & 
    21 ms& mAP: 81.5\%& $\checkmark$ &   &     \\ \hline
   
    \rowcolor{LightGreen}
    Greer et al. & 2023 & \cite{greer2023robust} &  \colorbox{LightGreen}{\makecell[l]{Deformable DETR with \\ custom salient-light loss}}  & LAVA~\cite{greer2023lava} & N/A & N/A & $\checkmark$ & $\checkmark$ &   \\ \hline

    \end{tabular}
}
\end{table*}

\subsection{Modifications of Generic Object Detectors}

The first group comprises approaches that use an existing CNN-based model for generic object detection with minor modifications to compensate for smaller object sizes. The corresponding approaches are marked green in Table \ref{tab:approaches}. Generic object detectors are especially favorable due to their inference speed.

The earliest approach to modify YOLO~\cite{redmon2016you} for the TLR task was presented by Jensen et al.~\cite{jensen2017evaluating}. Here, YOLOv2~\cite{redmon2018yolo9000} was modified by removing the last convolutional layer and adding three $3\times 3$ convolutional layers with 1024 filters, followed by a $1\times1$ convolutional layer with the number of outputs needed for the specific detection. This model, however, only performed detection, not the classification of the TL states. 
Bali et al.~\cite{bali2022lightweight}  tried to replace the YOLOv2 backbone with different lightweight CNNs, whereas the best results were achieved with SqueezeNet \cite{iandola2016squeezenet}.

Müller and Dietmayer~\cite{muller2018detecting} presented a modified version of the SSD~\cite{liu2016ssd} architecture for TLR with Inception-v3 \cite{szegedy2016rethinking} instead of VGG \cite{simonyan2015very} backbone for a better accuracy-speed trade-off. The authors analyzed the layer and feature map sizes of Inception-v3 and showed that they cannot guarantee the detection of objects with a width of 5 pixels. Therefore, to increase the recall on small objects, they introduced modified priors placed not in the center of each feature cell but arbitrarily using the offset vectors. Furthermore, early and late feature layers were concatenated for the BBox and confidence prediction to use context information from the early layers better. As in SSD, the confidence loss was formulated as a two-class problem (TL vs. background). Also a further layer was added to detect the TL state (\textit{red, yellow, green, off}). 

Faster R-CNN~\cite{ren2015faster} was first applied by Pon et al. \cite{pon2018hierarchical} for TLR within the joint traffic light and traffic sign detection network. Bach et al. \cite{bach2018deep} suggested further modifications to Faster R-CNN for TLR. In particular, some layers of the feature extractor networks (ResNet-50) were modified. Furthermore, anchors were determined not arbitrarily but via k-means clustering of the training set BBoxes. Finally, the loss function was expanded to allow for TL classification. Han et al.~\cite{han2019real} used the modified Faster R-CNN with VGG16 backbone for traffic sign and traffic light detection. To account for small object size, a small region proposal generator was used. For this, the \textit{pool4} layer of VGG16 was removed. Additionally, the online hard examples mining (OHEM)~\cite{shrivastava2016training} approach was applied to locate small objects more robustly and helped to increase mAP by 2-3 pp. The best results, however, were achieved with ResNet-50 \cite{he2016deep} with dilation.

Abraham et al.~\cite{abraham2021traffic} used a modified YOLOv4~\cite{bochkovskiy2020yolov4} with cross-stage partial connections (CSP). The feature extractor contained a Darknet53 \cite{redmon2018yolov3} backbone, a path aggregation network, spatial pyramid pooling, and a spatial attention module, while the detector used the YOLOv4 head. A similar approach was followed by Wang et al. \cite{wang2022traffic}. Here, YOLOv4 with  CSPDarknet-53 feature extraction network was modified by fusing certain layers and enhancing the shallow features. Furthermore, the BBox uncertainty prediction was also added. Lastly, Zhao et al. \cite{zhao2022study} showed that ShuffleNet \cite{zhang2018shufflenet} leads to better results when used as a backbone in YOLOv4.

The work by Ennahhal et al. \cite{ennahhal2019real} is one of the few that compared several approaches. Their results show that Faster R-CNN outperformed R-FCN~\cite{dai2016rfcnLHS16} and SSD in terms of mAP. Later, Gokul et al. \cite{gokul2020comparative} have also demonstrated that Faster-R-CNN has the best trade-off between accuracy and speed compared to YOLOv2 and YOLOv3.

Liu and Li~\cite{liu2023traffic} proposed to modify the backbone of the YOLOv5\footnote{https://github.com/ultralytics/yolov5}. The custom backbone architecture is inspired by the U2Net~\cite{qin2020u2net} and contains a series of residual U-blocks. Additionally, the authors replace the C3 modules in the neck part of YOLOv5 with ConvNextBlocks~\cite{liu2022aconvnet} to improve feature extraction. The resulting model has demonstrated better accuracy compared to the baseline YOLOv5. Models based on YOLOv5s have demonstrated a remarkable inference speed of 48 FPS.

Finally, a single approach that goes beyond CNN-based object detectors is that by Greer et al.~\cite{greer2023robust}. The authors used the Deformable DETR~\cite{zhu2021deformable}, a generic object detector with a transformer encoder-decoder architecture and features extracted using a CNN backbone (ResNet-50). The authors evaluated the impact of the salience-sensitive focal loss and showed better performance on salient traffic lights.

\subsection{Multi-Stage Approaches}

The second group contains approaches where the TLR task is split into two subtasks: detection and classification, s.t. a separate model is used for each of them. The corresponding approaches are marked blue in Table \ref{tab:approaches}.

\textbf{Generic object detector +  CNN for classification:} In the work by Behrendt et al.~\cite{behrendt2017deep} introducing the BSTLD dataset, YOLO was modified to detect TL objects as small as $3\times10$ pixels. For this, the authors took random crops of size $448\times448$ from an image. Also, the number of grid cells was increased from $7\times7$ to $11\times11$. The classification part of the original YOLO loss was removed. Instead, a small classification network consisting of three convolutional and three fully-connected layers was used to detect TL states. 

Lu et al.~\cite{lu2018traffic} proposed an approach consisting of two parts: the first one proposes attention regions that can contain traffic lights, and the second part performs localization and classifications on the cropped and resized attention regions found by the first model. Both blocks follow the Faster R-CNN architecture. 

A similar approach was followed by Wang et al.~\cite{wang2018method}, who used YOLOv3 \cite{redmon2018yolov3} for the detection of regions of interest (ROI). The classification of a TL status was performed with a lightweight CNN consisting of two convolutional and two max-pooling layers. Similarly, as in the previous work, the lightweight CNN gets ROIs from the YOLOv3 as input and predicts one of the four states (\textit{red, green, yellow, unknown}).

Cai et al. \cite{cai2019deltr} proposed a two-stage approach, where the detection part consisted of the SSDLite with MobileNetv2~\cite{sandler2018mobilenetv2}, whereas classification was performed by a small three-layer network.

In the work by Kim et al. \cite{kim2019traffic}, the detection stage is performed by a semantic segmentation network, which is then used to calculate BBoxes. This is motivated by its better performance on very small objects. In particular, a binary version of the ENet \cite{paszke2016enet} is used. For the classification part, a LeNet-5-based \cite{leCun1998gradient} model is used. This model was shown to beat Faster R-CNN from the previous work by authors \cite{kim2018efficient} both in terms of accuracy and speed.

Jayasinghe et al.~\cite{jayasinghe2022towards} used a two-stage approach, where detection was performed either with Faster R-CNN with a ResNet-50 backbone or SSD with MobileNetv2 backbone, and the classification was done with ResNet-18.

\textbf{Generic object detector + non-deep learning approach for classification:} 
Kim et al.~\cite{kim2018deep} used an unmodified SSD with a standard VGG16 backbone as a coarse-grained detector. The fine-grained detection is performed via spatiotemporal filtering and has the goal to compensate for the poor performance of SSD on small objects. The latter uses a point-based reward system; the points are rewarded for detections consistent in the spatial and temporal domains.

Yudin et al.~\cite{yudin2018usage} used a U-Net~\cite{ronneberger15unet}-inspired fully-convolutional network to predict a grayscale map of TL locations, which is further binarized using thresholding. After that, the detected regions are clustered using DBSCAN and filtered, yielding the predicted location. The proposed approach is shown to lead to higher precision and recall compared to the SSD300.

Gupta and Choudhary~\cite{gupta2019framework} used Grassman manifold learning for TL and pictogram classification, while the detection step was performed with a Faster R-CNN. For the TL classification, features extracted from VGG16 were used to create subspaces on a Grassman manifold for each TL state. After that, discriminant analysis on the manifold was used to distinguish between TLs.

In the work of Tran et al. \cite{tran2020accurate}, the detections and classifications made by YOLOv4 are additionally processed by a color-based clustering method to remove irrelevant predictions. Moreover, a rule-based heuristic to identify the most important TL in an input image is applied as the last step. Similarly, Nguyen et al. \cite{nguyen2020robust} validate the predictions done by YOLOv3 via hand-crafted features and classification using HSV color space.

\textbf{Non-deep learning detector + CNN for classification:}
Wang et al. \cite{wang2018traffic} used a high dynamic range camera to get input images for different channels; this allowed them to detect TL ROIs from input images using a saliency map. Then, a customized AlexNet was used for the TL classification. Kim et al. \cite{kim2018efficient} also used a color-based approach. They proposed transforming an input image to another color space before passing it to a generic object detector. Different models represented the latter, whereas Faster R-CNN with Inception-ResNet-v2 was shown to be the most suitable for the task. The HSV color space was used in work by Gao et al. \cite{gao2020hybrid} to generate the ROIs, whereas the classification was performed with AlexNet. Vitas et al. \cite{vitas2020traffic} applied adaptive thresholding to generate ROIs at the detection step, whereas the classification was done with a simple three-layer CNN.

\textbf{Further approaches:} Possatti et al.~\cite{possatti2019traffic} incorporated the usage of prior maps containing coordinates of TLs, whereas YOLOv3 was used for TLR. YOLOv3 was not additionally modified and trained to distinguish between two classes: \textit{red-yellow} and \textit{green} TLs. The TL position is projected to the image plane using the data from the prior maps and the vehicle localization data. Finally, only those BBoxes predicted by the YOLOv3 corresponding to the projected map objects are used for final predictions.

Yeh et al.~\cite{yeh2019traffic,yeh2021traffic} presented a three-stage approach, where YOLOv3 first localizes traffic lights. Next, YOLOv3-tiny detects the TL states. Finally, LeNet is applied to classify the arrows in different directions. HD maps and collected LiDAR data are used to find the TL position.

\subsection{Task-specific Single-stage Approaches}
Finally, the third group comprises those approaches where TLR is performed within a single network deliberately designed for this task. The corresponding methods are marked yellow in Table \ref{tab:approaches}. Unlike most methods, which follow the two-step approach involving TL detection and subsequent classification, the DeepTLR by Weber et al.~\cite{weber2016deeptlr} is a pure CNN that directly classifies each fine-grained pixel region over the image, thus creating a probability map for each of three classes: \textit{red, yellow,} and \textit{green}. For the pixels in probability maps, which surpass a certain threshold, BBox prediction is performed. The feature extraction part of DeepTLR uses the AlexNet architecture~\cite{krizhevsky2012imagenet}, whereas the BBox regression follows that of the OverFeat~\cite{sermanet2014overfeat}.


The HDTLR approach~\cite{weber2018hdtlr} by Weber et al. builds upon DeepTLR, extending and improving the detection part. Unlike DeepTLR, HDTLR can use any CNN for the feature extraction part. Experiments were performed with AlexNet, GoogLeNet, and VGG, while the latter performed the best.

Wang et al. \cite{wang2022simultaneous} proposed a joint detection and tracking approach, whereas a CNN and integrated channel feature tracking are used to predict both TL coordinates and states. 

\section{Conclusion}

In this paper, we gave an overview of the existing works on traffic light recognition. Our analysis has revealed that the predominant approach in the literature is the modification of a generic object detector like YOLO, SSD, or Faster R-CNN. In particular, YOLO versions 1-5 were used especially often. A large group of multi-stage approaches uses an existing detector as an attention or region proposal module, which determines the positions of the traffic lights, whereas an additional CNN classifier distinguishes between traffic light states and pictograms. This classification network usually has a very simple architecture. Less popular is the usage of a rule-based ROI detector or of a non-CNN classification method. Finally, a separate cluster of approaches is formed by methods that perform traffic light recognition within a single model so that the task is learned end-to-end without intrinsic separation into detection and classification steps. 


Furthermore, our overview has shown that a lot of works reach real-time performance, but perform evaluation on private datasets, which makes a fair comparison of different methods difficult. We also have determined that, unlike most object detection tasks, open-sourcing the code of the TLR models is still rare. We hope our findings facilitate further research on traffic light recognition.

\section*{ACKNOWLEDGMENT}

The research leading to these results is funded by the German Federal Ministry for Economic Affairs and Climate Action within the project “Shuttle2X“ (grant 19S22001B).

{\small
\bibliographystyle{IEEEtran}
\bibliography{references}

\begin{thebibliography}{10}
\providecommand{\url}[1]{#1}
\csname url@samestyle\endcsname
\providecommand{\newblock}{\relax}
\providecommand{\bibinfo}[2]{#2}
\providecommand{\BIBentrySTDinterwordspacing}{\spaceskip=0pt\relax}
\providecommand{\BIBentryALTinterwordstretchfactor}{4}
\providecommand{\BIBentryALTinterwordspacing}{\spaceskip=\fontdimen2\font plus
\BIBentryALTinterwordstretchfactor\fontdimen3\font minus
  \fontdimen4\font\relax}
\providecommand{\BIBforeignlanguage}[2]{{%
\expandafter\ifx\csname l@#1\endcsname\relax
\typeout{** WARNING: IEEEtran.bst: No hyphenation pattern has been}%
\typeout{** loaded for the language `#1'. Using the pattern for}%
\typeout{** the default language instead.}%
\else
\language=\csname l@#1\endcsname
\fi
#2}}
\providecommand{\BIBdecl}{\relax}
\BIBdecl

\bibitem{cabrera2015asurvey}
M.~D. Cabrera, P.~Cerri, G.~Pirlo, M.~A. Ferrer, and D.~Impedovo, ``A survey on
  traffic light detection,'' in \emph{New Trends in Image Analysis and
  Processing - {ICIAP} - Workshops}.\hskip 1em plus 0.5em minus 0.4em\relax
  Springer, 2015.

\bibitem{jensen2016vision}
M.~B. Jensen, M.~P. Philipsen, A.~M{\o}gelmose, T.~B. Moeslund, and M.~M.
  Trivedi, ``Vision for looking at traffic lights: Issues, survey, and
  perspectives,'' \emph{IEEE transactions on intelligent transportation
  systems}, 2016.

\bibitem{gautam2023image}
S.~Gautam and A.~Kumar, ``Image-based automatic traffic lights detection system
  for autonomous cars: a review,'' \emph{Multimedia Tools and Applications},
  2023.

\bibitem{muller2018detecting}
J.~M{\"u}ller and K.~Dietmayer, ``Detecting traffic lights by single shot
  detection,'' in \emph{International Conference on Intelligent Transportation
  Systems (ITSC)}.\hskip 1em plus 0.5em minus 0.4em\relax IEEE, 2018.

\bibitem{weber2018hdtlr}
M.~Weber, M.~Huber, and J.~M. Z{\"o}llner, ``Hdtlr: A cnn based hierarchical
  detector for traffic lights,'' in \emph{International Conference on
  Intelligent Transportation Systems (ITSC)}.\hskip 1em plus 0.5em minus
  0.4em\relax IEEE, 2018.

\bibitem{LARA}
``{Robotics Centre of Mines ParisTech. Traffic lights recognition (tlr) public
  benchmarks, 2015.}''

\bibitem{WPI}
Z.~Chen and X.~Huang, ``Accurate and reliable detection of traffic lights using
  multiclass learning and multiobject tracking,'' \emph{{IEEE} Intell. Transp.
  Syst. Mag.}, 2016.

\bibitem{behrendt2017deep}
K.~Behrendt, L.~Novak, and R.~Botros, ``A deep learning approach to traffic
  lights: Detection, tracking, and classification,'' in \emph{International
  Conference on Robotics and Automation (ICRA)}.\hskip 1em plus 0.5em minus
  0.4em\relax IEEE, 2017.

\bibitem{fregin2018driveu}
A.~Fregin, J.~Muller, U.~Krebel, and K.~Dietmayer, ``The driveu traffic light
  dataset: Introduction and comparison with existing datasets,'' in
  \emph{International Conference on Robotics and Automation (ICRA)}.\hskip 1em
  plus 0.5em minus 0.4em\relax IEEE, 2018.

\bibitem{janosovits2022cityscapes}
J.~Janosovits, ``Cityscapes tl++: Semantic traffic light annotations for the
  cityscapes dataset,'' in \emph{International Conference on Robotics and
  Automation (ICRA)}.\hskip 1em plus 0.5em minus 0.4em\relax IEEE, 2022.

\bibitem{yang2022scrdet++}
X.~Yang, J.~Yan, W.~Liao, X.~Yang, J.~Tang, and T.~He, ``Scrdet++: Detecting
  small, cluttered and rotated objects via instance-level feature denoising and
  rotation loss smoothing,'' \emph{IEEE Transactions on Pattern Analysis and
  Machine Intelligence}, 2022.

\bibitem{COCO}
{David Kirchhoff and Philip Hoang}, ``{COCO Dataset Extensions for Driving
  Tasks},'' \url{https://www.neuralception.com/cocodatasetextension/},
  accessed: 2023-04-01.

\bibitem{lin2014microsoft}
T.-Y. Lin, M.~Maire, S.~Belongie, J.~Hays, P.~Perona, D.~Ramanan,
  P.~Doll{\'a}r, and C.~L. Zitnick, ``Microsoft coco: Common objects in
  context,'' in \emph{European Conference on Computer Vision (ECCV)}.\hskip 1em
  plus 0.5em minus 0.4em\relax Springer, 2014.

\bibitem{Cordts2016Cityscapes}
M.~Cordts, M.~Omran, S.~Ramos, T.~Rehfeld, M.~Enzweiler, R.~Benenson,
  U.~Franke, S.~Roth, and B.~Schiele, ``The cityscapes dataset for semantic
  urban scene understanding,'' in \emph{Conference on Computer Vision and
  Pattern Recognition (CVPR)}, 2016.

\bibitem{Roboflow}
{Roboflow}, ``{Self Driving Car Dataset},''
  \url{https://public.roboflow.com/object-detection/self-driving-car}, 2021,
  accessed: 2023-03-01.

\bibitem{Udacity}
{Udacity}, ``{self-driving-car},''
  \url{https://github.com/udacity/self-driving-car}, 2021, accessed:
  2023-03-01.

\bibitem{Waymo}
{Waymo}, ``{Waymo Open Dataset},'' \url{https://waymo.com/open/}, accessed:
  2023-03-01.

\bibitem{BDD100k}
F.~Yu, H.~Chen, X.~Wang, W.~Xian, Y.~Chen, F.~Liu, V.~Madhavan, and T.~Darrell,
  ``{BDD100K:} {A} diverse driving dataset for heterogeneous multitask
  learning,'' in \emph{Conference on Computer Vision and Pattern Recognition
  (CVPR)}, 2020.

\bibitem{ApolloScape}
{ApolloScape Website}, ``{ApolloScape Dataset},''
  \url{https://apolloscape.auto/}, accessed: 2023-03-01.

\bibitem{gautam2022automatic}
S.~Gautam and A.~Kumar, ``Automatic traffic light detection for self-driving
  cars using transfer learning,'' in \emph{Intelligent Sustainable Systems:
  Selected Papers of WorldS4}.\hskip 1em plus 0.5em minus 0.4em\relax Springer,
  2022.

\bibitem{john2014traffic}
V.~John, K.~Yoneda, B.~Qi, Z.~Liu, and S.~Mita, ``Traffic light recognition in
  varying illumination using deep learning and saliency map,'' in
  \emph{International Conference on Intelligent Transportation Systems
  (ITSC)}.\hskip 1em plus 0.5em minus 0.4em\relax {IEEE}, 2014.

\bibitem{john2015saliency}
V.~John, K.~Yoneda, Z.~Liu, and S.~Mita, ``Saliency map generation by the
  convolutional neural network for real-time traffic light detection using
  template matching,'' \emph{{IEEE} Trans. Computational Imaging}, 2015.

\bibitem{weber2016deeptlr}
M.~Weber, P.~Wolf, and J.~M. Z{\"o}llner, ``Deeptlr: A single deep
  convolutional network for detection and classification of traffic lights,''
  in \emph{Intelligent Vehicles Symposium (IV)}.\hskip 1em plus 0.5em minus
  0.4em\relax IEEE, 2016.

\bibitem{jensen2017evaluating}
M.~B. Jensen, K.~Nasrollahi, and T.~B. Moeslund, ``Evaluating state-of-the-art
  object detector on challenging traffic light data,'' in \emph{Conference on
  Computer Vision and Pattern Recognition (CVPR) - Workshops}, 2017.

\bibitem{pon2018hierarchical}
A.~Pon, O.~Adrienko, A.~Harakeh, and S.~L. Waslander, ``A hierarchical deep
  architecture and mini-batch selection method for joint traffic sign and light
  detection,'' in \emph{Conference on Computer and Robot Vision (CRV)}.\hskip
  1em plus 0.5em minus 0.4em\relax IEEE, 2018.

\bibitem{bach2018deep}
M.~Bach, D.~Stumper, and K.~Dietmayer, ``Deep convolutional traffic light
  recognition for automated driving,'' in \emph{International Conference on
  Intelligent Transportation Systems (ITSC)}.\hskip 1em plus 0.5em minus
  0.4em\relax IEEE, 2018.

\bibitem{kim2018efficient}
H.-K. Kim, J.~H. Park, and H.-Y. Jung, ``An efficient color space for
  deep-learning based traffic light recognition,'' \emph{Journal of Advanced
  Transportation}, vol. 2018, 2018.

\bibitem{lu2018traffic}
Y.~Lu, J.~Lu, S.~Zhang, and P.~Hall, ``Traffic signal detection and
  classification in street views using an attention model,''
  \emph{Computational Visual Media}, 2018.

\bibitem{wang2018traffic}
J.-G. Wang and L.-B. Zhou, ``Traffic light recognition with high dynamic range
  imaging and deep learning,'' \emph{IEEE Transactions on Intelligent
  Transportation Systems}, 2018.

\bibitem{kim2018deep}
J.~Kim, H.~Cho, M.~Hwangbo, J.~Choi, J.~Canny, and Y.~P. Kwon, ``Deep traffic
  light detection for self-driving cars from a large-scale dataset,'' in
  \emph{International Conference on Intelligent Transportation Systems
  (ITSC)}.\hskip 1em plus 0.5em minus 0.4em\relax IEEE, 2018.

\bibitem{wang2018method}
X.~Wang, T.~Jiang, and Y.~Xie, ``A method of traffic light status recognition
  based on deep learning,'' in \emph{International Conference on Robotics,
  Control and Automation Engineering}, 2018.

\bibitem{yudin2018usage}
D.~Yudin and D.~Slavioglo, ``Usage of fully convolutional network with
  clustering for traffic light detection,'' in \emph{Mediterranean Conference
  on Embedded Computing (MECO)}.\hskip 1em plus 0.5em minus 0.4em\relax IEEE,
  2018.

\bibitem{han2019real}
C.~Han, G.~Gao, and Y.~Zhang, ``Real-time small traffic sign detection with
  revised faster-rcnn,'' \emph{Multimedia Tools and Applications}, 2019.

\bibitem{possatti2019traffic}
L.~C. Possatti, R.~Guidolini, V.~B. Cardoso, R.~F. Berriel, T.~M. Paix{\~a}o,
  C.~Badue, A.~F. De~Souza, and T.~Oliveira-Santos, ``Traffic light recognition
  using deep learning and prior maps for autonomous cars,'' in
  \emph{International Joint Conference on Neural Networks (IJCNN)}.\hskip 1em
  plus 0.5em minus 0.4em\relax IEEE, 2019.

\bibitem{ennahhal2019real}
Z.~Ennahhal, I.~Berrada, and K.~Fardousse, ``Real time traffic light detection
  and classification using deep learning,'' in \emph{International Conference
  on Wireless Networks and Mobile Communications (WINCOM)}.\hskip 1em plus
  0.5em minus 0.4em\relax IEEE, 2019.

\bibitem{gupta2019framework}
A.~Gupta and A.~Choudhary, ``A framework for traffic light detection and
  recognition using deep learning and grassmann manifolds,'' in
  \emph{Intelligent Vehicles Symposium (IV)}.\hskip 1em plus 0.5em minus
  0.4em\relax IEEE, 2019.

\bibitem{du2019real}
L.~Du, W.~Chen, S.~Fu, H.~Kong, C.~Li, and Z.~Pei, ``Real-time detection of
  vehicle and traffic light for intelligent and connected vehicles based on
  yolov3 network,'' in \emph{International Conference on Transportation
  Information and Safety (ICTIS)}.\hskip 1em plus 0.5em minus 0.4em\relax IEEE,
  2019.

\bibitem{yeh2019traffic}
T.-W. Yeh, S.-Y. Lin, H.-Y. Lin, S.-W. Chan, C.-T. Lin, and Y.-Y. Lin,
  ``Traffic light detection using convolutional neural networks and lidar
  data,'' in \emph{International Symposium on Intelligent Signal Processing and
  Communication Systems (ISPACS)}.\hskip 1em plus 0.5em minus 0.4em\relax IEEE,
  2019.

\bibitem{yeh2021traffic}
T.-W. Yeh, H.-Y. Lin, and C.-C. Chang, ``Traffic light and arrow signal
  recognition based on a unified network,'' \emph{Applied Sciences}, 2021.

\bibitem{kim2019traffic}
H.-K. Kim, K.-Y. Yoo, J.~H. Park, and H.-Y. Jung, ``Traffic light recognition
  based on binary semantic segmentation network,'' \emph{Sensors}, 2019.

\bibitem{aneesh2019real}
A.~Aneesh, L.~Shine, R.~Pradeep, and V.~Sajith, ``Real-time traffic light
  detection and recognition based on deep retinanet for self driving cars,'' in
  \emph{International Conference on Intelligent Computing, Instrumentation and
  Control Technologies (ICICICT)}.\hskip 1em plus 0.5em minus 0.4em\relax IEEE,
  2019.

\bibitem{vishal2019traffic}
K.~Vishal, C.~Arvind, R.~Mishra, and V.~Gundimeda, ``Traffic light recognition
  for autonomous vehicles by admixing the traditional ml and dl,'' in
  \emph{International Conference on Machine Vision (ICMV)}.\hskip 1em plus
  0.5em minus 0.4em\relax SPIE, 2019.

\bibitem{cai2019deltr}
Y.~Cai, C.~Li, S.~Wang, and J.~Cheng, ``Deltr: A deep learning based approach
  to traffic light recognition,'' in \emph{Image and Graphics: 10th
  International Conference, ICIG}.\hskip 1em plus 0.5em minus 0.4em\relax
  Springer, 2019.

\bibitem{janahiraman2019traffic}
T.~V. Janahiraman and M.~S.~M. Subuhan, ``Traffic light detection using
  tensorflow object detection framework,'' in \emph{International Conference on
  System Engineering and Technology (ICSET)}.\hskip 1em plus 0.5em minus
  0.4em\relax IEEE, 2019.

\bibitem{ouyang2020deep}
Z.~Ouyang, J.~Niu, Y.~Liu, and M.~Guizani, ``Deep cnn-based real-time traffic
  light detector for self-driving vehicles,'' \emph{{IEEE} Trans. Mob.
  Comput.}, 2020.

\bibitem{tran2020accurate}
T.~H.-P. Tran and J.~W. Jeon, ``Accurate real-time traffic light detection
  using yolov4,'' in \emph{IEEE International Conference on Consumer
  Electronics-Asia (ICCE-Asia)}.\hskip 1em plus 0.5em minus 0.4em\relax IEEE,
  2020.

\bibitem{nguyen2020robust}
P.~M. Nguyen, V.~C. Nguyen, S.~N. Nguyen, L.~M.~T. Dang, H.~X. Nguyen, and
  V.~D. Nguyen, ``Robust traffic light detection and classification under day
  and night conditions,'' in \emph{International conference on control,
  automation and systems (ICCAS)}.\hskip 1em plus 0.5em minus 0.4em\relax IEEE,
  2020.

\bibitem{nguyen2019realtime}
V.~D. Nguyen, T.~D. Tran, J.~Y. Byun, and J.~W. Jeon, ``Real-time vehicle
  detection using an effective region proposal-based depth and 3-channel
  pattern,'' \emph{{IEEE} Trans. Intell. Transp. Syst.}, 2019.

\bibitem{gao2020hybrid}
F.~Gao and C.~Wang, ``Hybrid strategy for traffic light detection by combining
  classical and self-learning detectors,'' \emph{IET Intelligent Transport
  Systems}, 2020.

\bibitem{vitas2020traffic}
D.~Vitas, M.~Tomic, and M.~Burul, ``Traffic light detection in autonomous
  driving systems,'' \emph{IEEE Consumer Electronics Magazine}, 2020.

\bibitem{gokul2020comparative}
R.~Gokul, A.~Nirmal, K.~Bharath, M.~Pranesh, and R.~Karthika, ``A comparative
  study between state-of-the-art object detectors for traffic light
  detection,'' in \emph{International Conference on Emerging Trends in
  Information Technology and Engineering (ic-ETITE)}.\hskip 1em plus 0.5em
  minus 0.4em\relax IEEE, 2020.

\bibitem{abraham2021traffic}
A.~Abraham, D.~Purwanto, and H.~Kusuma, ``Traffic lights and traffic signs
  detection system using modified you only look once,'' in \emph{International
  Seminar on Intelligent Technology and Its Applications (ISITIA)}.\hskip 1em
  plus 0.5em minus 0.4em\relax IEEE, 2021.

\bibitem{yan2021end}
S.~Yan, X.~Liu, W.~Qian, and Q.~Chen, ``An end-to-end traffic light detection
  algorithm based on deep learning,'' in \emph{International conference on
  security, pattern analysis, and cybernetics (SPAC)}.\hskip 1em plus 0.5em
  minus 0.4em\relax IEEE, 2021.

\bibitem{xiang2021real}
N.~Xiang, Z.~Cao, Y.~Wang, and Q.~Jia, ``A real-time vehicle traffic light
  detection algorithm based on modified yolov3,'' in \emph{International
  Conference on Electronics Technology (ICET)}.\hskip 1em plus 0.5em minus
  0.4em\relax IEEE, 2021.

\bibitem{naimi2021fast}
H.~Naimi, T.~Akilan, and M.~A. Khalid, ``Fast traffic sign and light detection
  using deep learning for automotive applications,'' in \emph{IEEE Western New
  York Image and Signal Processing Workshop (WNYISPW)}.\hskip 1em plus 0.5em
  minus 0.4em\relax IEEE, 2021.

\bibitem{wang2022traffic}
Q.~Wang, Q.~Zhang, X.~Liang, Y.~Wang, C.~Zhou, and V.~I. Mikulovich, ``Traffic
  lights detection and recognition method based on the improved yolov4
  algorithm,'' \emph{Sensors}, 2022.

\bibitem{zhao2022study}
Y.~Zhao, Y.~Feng, Y.~Wang, Z.~Zhang, and Z.~Zhang, ``Study on detection and
  recognition of traffic lights based on improved yolov4,'' \emph{Sensors},
  2022.

\bibitem{bali2022lightweight}
S.~Bali, T.~Kumar, and S.~Tyagi, ``Lightweight deep learning model for traffic
  light detection,'' in \emph{International Conference on Technological
  Advancements in Computational Sciences (ICTACS)}.\hskip 1em plus 0.5em minus
  0.4em\relax IEEE, 2022.

\bibitem{wang2022simultaneous}
K.~Wang, X.~Tang, S.~Zhao, and Y.~Zhou, ``Simultaneous detection and tracking
  using deep learning and integrated channel feature for ambint traffic light
  recognition,'' \emph{Journal of Ambient Intelligence and Humanized
  Computing}, 2022.

\bibitem{jayasinghe2022towards}
O.~Jayasinghe, S.~Hemachandra, D.~Anhettigama, S.~Kariyawasam,
  T.~Wickremasinghe, C.~Ekanayake, R.~Rodrigo, and P.~Jayasekara, ``Towards
  real-time traffic sign and traffic light detection on embedded systems,'' in
  \emph{Intelligent Vehicles Symposium (IV)}.\hskip 1em plus 0.5em minus
  0.4em\relax IEEE, 2022.

\bibitem{mostafa2022yolo}
M.~Mostafa and M.~Ghantous, ``A yolo based approach for traffic light
  recognition for adas systems,'' in \emph{International Mobile, Intelligent,
  and Ubiquitous Computing Conference (MIUCC)}.\hskip 1em plus 0.5em minus
  0.4em\relax IEEE, 2022.

\bibitem{lin2022two}
S.-Y. Lin and H.-Y. Lin, ``A two-stage framework for diverse traffic light
  recognition based on individual signal detection,'' in \emph{Pattern
  Recognition and Artificial Intelligence: Mediterranean Conference
  (MedPRAI)}.\hskip 1em plus 0.5em minus 0.4em\relax Springer, 2022.

\bibitem{derong2023remote}
M.~DeRong and T.~ZhongMei, ``Remote traffic light detection and recognition
  based on deep learning,'' in \emph{World Conference on Computing and
  Communication Technologies (WCCCT)}.\hskip 1em plus 0.5em minus 0.4em\relax
  IEEE, 2023.

\bibitem{liu2023traffic}
P.~Liu and T.~Li, ``Traffic light detection based on depth improved yolov5,''
  in \emph{International Conference on Neural Networks, Information and
  Communication Engineering (NNICE)}.\hskip 1em plus 0.5em minus 0.4em\relax
  IEEE, 2023.

\bibitem{greer2023robust}
R.~Greer, A.~Gopalkrishnan, J.~Landgren, L.~Rakla, A.~Gopalan, and M.~M.
  Trivedi, ``Robust traffic light detection using salience-sensitive loss:
  Computational framework and evaluations,'' \emph{CoRR}, vol. abs/2305.04516,
  2023.

\bibitem{greer2023lava}
R.~Greer, A.~Gopalkrishnan, N.~Deo, A.~Rangesh, and M.~M. Trivedi, ``Salient
  sign detection in safe autonomous driving: {AI} which reasons over full
  visual context,'' \emph{CoRR}, 2023.

\bibitem{redmon2016you}
J.~Redmon, S.~Divvala, R.~Girshick, and A.~Farhadi, ``You only look once:
  Unified, real-time object detection,'' in \emph{Conference on Computer Vision
  and Pattern Recognition (CVPR)}, 2016.

\bibitem{redmon2018yolo9000}
J.~Redmon and A.~Farhadi, ``{YOLO9000:} better, faster, stronger,'' in
  \emph{Conference on Computer Vision and Pattern Recognition (CVPR)}.\hskip
  1em plus 0.5em minus 0.4em\relax {IEEE} Computer Society, 2017.

\bibitem{iandola2016squeezenet}
F.~N. Iandola, M.~W. Moskewicz, K.~Ashraf, S.~Han, W.~J. Dally, and K.~Keutzer,
  ``Squeezenet: Alexnet-level accuracy with 50x fewer parameters and
  {\textless}1mb model size,'' \emph{CoRR}, vol. abs/1602.07360, 2016.

\bibitem{liu2016ssd}
W.~Liu, D.~Anguelov, D.~Erhan, C.~Szegedy, S.~Reed, C.-Y. Fu, and A.~C. Berg,
  ``Ssd: Single shot multibox detector,'' in \emph{European Conference on
  Computer Vision (ECCV)}.\hskip 1em plus 0.5em minus 0.4em\relax Springer,
  2016.

\bibitem{szegedy2016rethinking}
C.~Szegedy, V.~Vanhoucke, S.~Ioffe, J.~Shlens, and Z.~Wojna, ``Rethinking the
  inception architecture for computer vision,'' in \emph{Conference on Computer
  Vision and Pattern Recognition (CVPR)}, 2016.

\bibitem{simonyan2015very}
K.~Simonyan and A.~Zisserman, ``Very deep convolutional networks for
  large-scale image recognition,'' in \emph{International Conference on
  Learning Representations (ICLR)}, 2015.

\bibitem{ren2015faster}
S.~Ren, K.~He, R.~B. Girshick, and J.~Sun, ``Faster {R-CNN:} towards real-time
  object detection with region proposal networks,'' in \emph{Advances in Neural
  Information Processing Systems (NIPS)}, 2015.

\bibitem{shrivastava2016training}
A.~Shrivastava, A.~Gupta, and R.~B. Girshick, ``Training region-based object
  detectors with online hard example mining,'' in \emph{Conference on Computer
  Vision and Pattern Recognition (CVPR)}.\hskip 1em plus 0.5em minus
  0.4em\relax {IEEE} Computer Society, 2016.

\bibitem{he2016deep}
K.~He, X.~Zhang, S.~Ren, and J.~Sun, ``Deep residual learning for image
  recognition,'' in \emph{Conference on Computer Vision and Pattern Recognition
  (CVPR)}.\hskip 1em plus 0.5em minus 0.4em\relax {IEEE} Computer Society,
  2016.

\bibitem{bochkovskiy2020yolov4}
A.~Bochkovskiy, C.~Wang, and H.~M. Liao, ``Yolov4: Optimal speed and accuracy
  of object detection,'' \emph{CoRR}, vol. abs/2004.10934, 2020.

\bibitem{redmon2018yolov3}
J.~Redmon and A.~Farhadi, ``Yolov3: An incremental improvement,'' \emph{CoRR},
  vol. abs/1804.02767, 2018.

\bibitem{zhang2018shufflenet}
X.~Zhang, X.~Zhou, M.~Lin, and J.~Sun, ``Shufflenet: An extremely efficient
  convolutional neural network for mobile devices,'' in \emph{Conference on
  Computer Vision and Pattern Recognition (CVPR)}.\hskip 1em plus 0.5em minus
  0.4em\relax Computer Vision Foundation / {IEEE} Computer Society, 2018.

\bibitem{dai2016rfcnLHS16}
J.~Dai, Y.~Li, K.~He, and J.~Sun, ``{R-FCN:} object detection via region-based
  fully convolutional networks,'' in \emph{Advances in Neural Information
  Processing Systems (NIPS)}, 2016.

\bibitem{qin2020u2net}
X.~Qin, Z.~Zhang, C.~Huang, M.~Dehghan, O.~R. Za{\"{\i}}ane, and
  M.~J{\"{a}}gersand, ``U\({}^{\mbox{2}}\)-net: Going deeper with nested
  u-structure for salient object detection,'' \emph{Pattern Recognition}, 2020.

\bibitem{liu2022aconvnet}
Z.~Liu, H.~Mao, C.~Wu, C.~Feichtenhofer, T.~Darrell, and S.~Xie, ``A convnet
  for the 2020s,'' in \emph{Conference on Computer Vision and Pattern
  Recognition (CVPR)}.\hskip 1em plus 0.5em minus 0.4em\relax {IEEE}, 2022.

\bibitem{zhu2021deformable}
X.~Zhu, W.~Su, L.~Lu, B.~Li, X.~Wang, and J.~Dai, ``Deformable {DETR:}
  deformable transformers for end-to-end object detection,'' in
  \emph{International Conference on Learning Representations (ICLR)}.\hskip 1em
  plus 0.5em minus 0.4em\relax OpenReview.net, 2021.

\bibitem{sandler2018mobilenetv2}
M.~Sandler, A.~Howard, M.~Zhu, A.~Zhmoginov, and L.-C. Chen, ``Mobilenetv2:
  Inverted residuals and linear bottlenecks,'' in \emph{Conference on Computer
  Vision and Pattern Recognition (CVPR)}, 2018.

\bibitem{paszke2016enet}
A.~Paszke, A.~Chaurasia, S.~Kim, and E.~Culurciello, ``Enet: {A} deep neural
  network architecture for real-time semantic segmentation,'' \emph{CoRR}, vol.
  abs/1606.02147, 2016.

\bibitem{leCun1998gradient}
Y.~LeCun, L.~Bottou, Y.~Bengio, and P.~Haffner, ``Gradient-based learning
  applied to document recognition,'' \emph{Proc. {IEEE}}, 1998.

\bibitem{ronneberger15unet}
O.~Ronneberger, P.~Fischer, and T.~Brox, ``U-net: Convolutional networks for
  biomedical image segmentation,'' in \emph{Medical Image Computing and
  Computer-Assisted Intervention - {MICCAI}}, ser. Lecture Notes in Computer
  Science.\hskip 1em plus 0.5em minus 0.4em\relax Springer, 2015.

\bibitem{krizhevsky2012imagenet}
A.~Krizhevsky, I.~Sutskever, and G.~E. Hinton, ``Imagenet classification with
  deep convolutional neural networks,'' in \emph{Advances in Neural Information
  Processing Systems (NIPS)}.

\bibitem{sermanet2014overfeat}
P.~Sermanet, D.~Eigen, X.~Zhang, M.~Mathieu, R.~Fergus, and Y.~LeCun,
  ``Overfeat: Integrated recognition, localization and detection using
  convolutional networks,'' in \emph{International Conference on Learning
  Representations (ICLR)}, 2014.

\end{thebibliography}
}

\end{document}